\documentclass[a4paper]{svproc}
\usepackage{graphicx, wrapfig, subcaption, caption}
\usepackage{amsmath, amssymb, amsfonts, siunitx}
\usepackage{mathtools}
\usepackage[dvipsnames]{xcolor}
\usepackage[hang,flushmargin]{footmisc}

\usepackage{url}

\DeclarePairedDelimiterX{\norm}[1]{\lVert}{\rVert}{#1}
\BeforeBeginEnvironment{wrapfigure}{\setlength{\intextsep}{0pt}}

\newcommand{\Alpha}{\textit{Alpha}}
\newcommand{\Bravo}{\textit{Bravo}}
\newcommand{\Al}{\mathcal{A}}
\newcommand{\Br}{\mathcal{B}}
\newcommand{\proj}[1]{\text{proj}_{#1}}
\newcommand{\R}[1]{\mathbb{R}^{#1}}
\newcommand{\SO}[1]{\textsf{SO}({#1})}
\newcommand{\SE}[1]{\textsf{SE}({#1})}
\newcommand{\vectorvar}[1]{\Vec{#1}}
\newcommand{\iden}[2]{\mathbb{I}_{#1\times#2}}
\newcommand{\zero}[2]{0_{#1\times#2}}
\newcommand{\fext}{\vectorvar{f}_{\mathrm{ext}}}

\begin{document}
\mainmatter              % start of a contribution
\title{Docking Multirotors in Close Proximity using Learnt Downwash Models}
\titlerunning{Close Proximity Docking using Learnt Downwash Models}  % abbreviated title (for running head)
\author{Ajay Shankar
\and Heedo Woo
\and Amanda Prorok
}
\authorrunning{Ajay Shankar et al.} % abbreviated author list (for running head)
\tocauthor{Ajay Shankar, Heedo Woo, Amanda Prorok}
\institute{Department of Computer Science \& Technology, University of Cambridge, UK\\
\email{\{as3233, hw527, asp45\}@cl.cam.ac.uk}
}

\maketitle              % typeset the title of the contribution

\begin{abstract}
Unmodeled aerodynamic disturbances pose a key challenge for multirotor flight when multiple vehicles are in close proximity to each other.
However, certain missions \textit{require} two multirotors to approach each other within 1-2 body-lengths of each other and hold formation --
we consider one such practical instance:
vertically docking two multirotors in the air.
In this leader-follower setting, the follower experiences significant downwash interference from the leader in its final docking stages.
To compensate for this, we employ a
learnt downwash model online within an optimal feedback controller to accurately track a docking maneuver and then hold formation.
Through real-world flights with different maneuvers, we demonstrate that this compensation is crucial for reducing the large vertical separation otherwise required by conventional/naive approaches.
Our evaluations show a tracking error of less than \SI{0.06}{m} for the follower (a 3-4x reduction) when approaching vertically within two body-lengths of the leader.
Finally, we deploy the complete system to effect a successful physical docking between two airborne multirotors in a single smooth planned trajectory.
\keywords{Multirotors, Aerial Docking, Geometric Learning}
\end{abstract}

\section{Motivation}
Multirotors are capable of executing agile maneuvers while retaining accurate tracking performance, making them well-suited for several applications centered around rapid-response, precise spatial monitoring and inspection tasks~\cite{chung2018survey}.
One practical use-case for such maneuverability is in enabling a multirotor aircraft to dock in-air with another.
This would allow enhanced long-range missions with multiple (potentially even heterogeneous) aircraft, where a larger `leader' aircraft tows and docks smaller `followers' that carry, for instance, more targeted sensors.
Prior work has shown successful demonstrations of docking unmanned aircraft~\cite{miyazaki2018airborne,wilson2015guidance}, albeit with some constraints (such as large separations).

\begin{figure*}
    \centering
    \begin{subfigure}[c]{0.32\textwidth}
        \includegraphics[width=\textwidth]{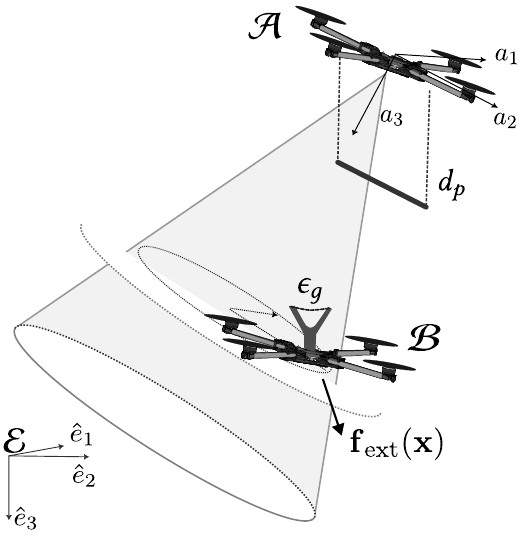}
    \end{subfigure}
    \begin{subfigure}[c]{0.28\textwidth}
        \includegraphics[width=\textwidth]{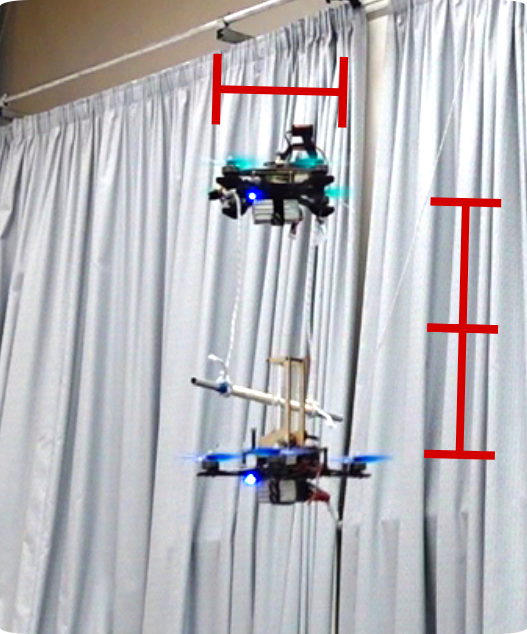}
    \end{subfigure}
    \begin{subfigure}[c]{0.31\textwidth}
        \includegraphics[width=\textwidth]{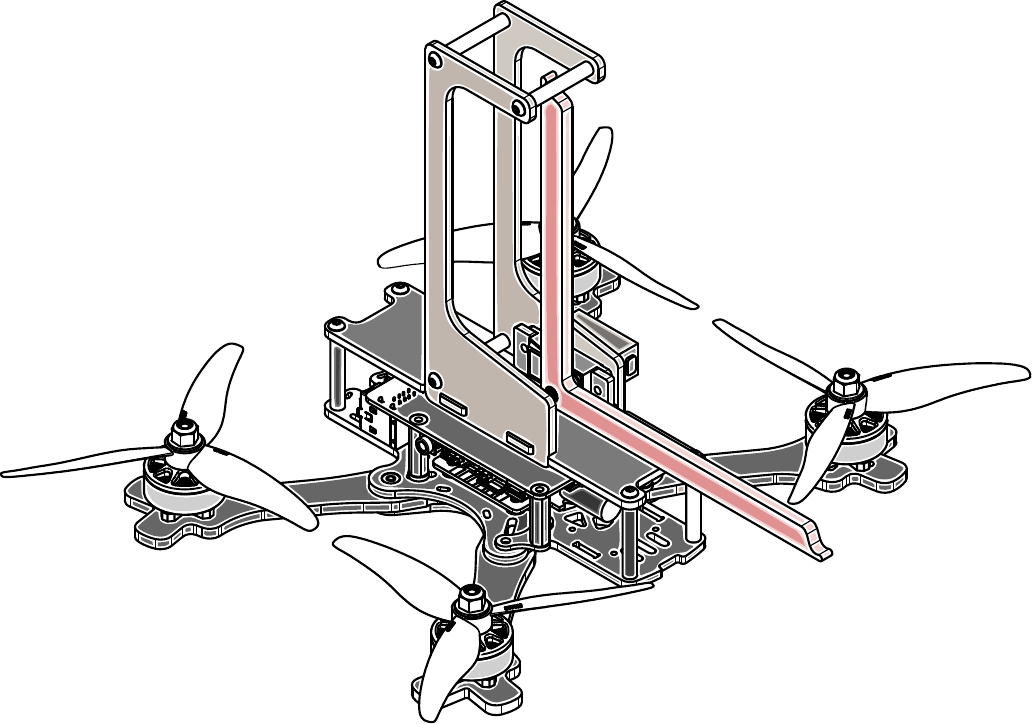}
    \end{subfigure}
    \caption{\textit{(Left)} An illustration of the two-vehicle docking problem, and the frames of reference used. \textit{(Middle)} Stills from our flights showing the two vehicles docked in close formation. \textit{(Right)} A CAD drawing of the servo-actuated docking mechanism (highlighted in red) mounted atop one of the vehicles.}
    \label{fig:illustration}
    %\vskip-4ex
\end{figure*}

A significant challenge for the agility, and indeed, the stability of these platforms is the presence of unmodeled exogenic effects, such as turbulent airflows from neighbors.
In some cases, a downwash-aware planner can circumvent some of these challenges.
However, this is completely unavoidable when designing missions that \textit{require} dense, close-proximity formation flights -- such as in docking.
In particular, consider a leader aircraft that tows a vertically suspended ``homing" platform for a follower to dock onto.
Naturally, since the follower must hover and keep formation directly underneath the leader at some point, it will experience strong downwash that can be hard for a disturbance observer to estimate online since they have a finite (non-zero) convergence time. 
The trivial solution of increasing their separation (4-5 body-lengths) poses additional risk from long suspended cables and underactuated dynamics.
Keeping the separation small (1-2 body-lengths), on the other hand, requires fast and accurate rejection of downwash disturbances, which our work considers.

In this work, we develop an efficient framework for \textit{close-proximity} docking maneuvers by compensating for downwash on the follower.
We build upon our prior work in modeling downwash patterns~\cite{smith2023so} by
representing it as a collective time-varying force, $\fext(t) \in \R{3}$, acting on the body, and then \textit{learning} it using state and control residuals from merely \SI{5}{mins} of real-world data.
This learnt model is then embedded online within an optimal planning and control framework running entirely onboard the follower, whose mission is to effect a docking maneuver with the leader's docking platform.
Our experiments show that a follower can remain with a tracking error of $\approx\SI{0.06}{m}$ and with negligible loss of performance even within 1-2 body-lengths underneath a leader.
We employ this model to our advantage in this work by wrapping its force predictions over a docking maneuver that requires the follower to be within this distance with a sufficiently small positioning error.

\textbf{Problem Statement.}
Consider two multirotors \Alpha{} ($\mathcal{A}$) and \Bravo{} ($\mathcal{B}$) designated `leader' and `follower' aircraft, respectively.
The leader flies a predetermined path, and carries a suspended docking platform displaced vertically at a distance, $d_p$. 
The follower has a mechanical gripper that triggers docking when positioned within a tolerance, $\epsilon_{g}$, of the platform.

The objective is to use an accurate model of $\fext$ and deploy it online so that
$|| \vectorvar{p}^{\mathcal{B}} - \vectorvar{p}_{des}^{\mathcal{B}} || \leq \epsilon_{g}$
for all $t$ within a time horizon $t_h$ when 
$|| \vectorvar{p}(t)^{\mathcal{A}} - \vectorvar{p}(t)^{\mathcal{B}} || < D$.
That is, we require \Bravo{} to maintain accurate positioning ($\vectorvar{p}$) while in close vertical proximity ($D$) of \Alpha{}.
For practicality, we also require the gripper's mechanism to admit active and passive triggers to reduce missed docking attempts.

\textbf{Related Work.}
Docking aircraft mid-flight has been investigated in different contexts (manned and unmanned).
Prior work in docking multirotors has used long suspended cables ($\approx\SI{4}{m}$) with winch mechanisms that limit the impact of downwash and swinging cables~\cite{miyazaki2018airborne} .
This can limit the leader's maneuverability and requires it to remain in hover.
Docking in the other direction (approaching from the top of the leader) can circumvent some of these challenges by employing `landing' strategies, since the leader's airflow is more laminar and the follower's downwash is reduced (assuming a smaller follower).
This has been demonstrated in the context of battery swapping~\cite{jain2020docking}, where a smaller docking vehicle settles exactly on top of the leader.

A key enabler for physical docking -- compensating for downwash -- has also been studied separately through analytical~\cite{jain2019modeling,zhang2020numerical} and data-driven~\cite{smith2023so,shi2020neural,li2023nonlinear} modeling.
However, large real-world robotics datasets are hard/expensive to obtain.
Consequently, we wish to maximize sample efficiency (a sample's contribution to the overall learning problem).
We achieve this by exploiting known symmetries and structure of the learnt function.
Imposing such inductive biases into the learnt function, especially via geometric priors, is an active robotics research domain~\cite{zhu2022sample}. 
Our prior work in learning multirotor downwash~\cite{smith2023so} has demonstrated dramatic boosts in training efficiency due entirely to our formulation that uses geometric equivariance over the \SO{2} group%
\footnote{
Please refer for more references \& detailed explanations of group invariance and equivariance.}.
Here, we apply that theory for the docking case, where accurate tracking close to \Alpha{} is a pass/fail criteria.

\section{Technical Approach}
We consider the multirotor $\mathcal{B}$ as a six degree-of-freedom object of mass \textit{m} with second-order non-linear dynamics in an inertial frame $\mathcal{E}$ evolving according to
$m \Vec{a} = -R^\mathcal{B}_\mathcal{E}\cdot T + mg\hat{e}_3,$
where $T$ is the collective thrust,
$R^\mathcal{B}_\mathcal{E}$
denotes the rotation between the fixed-inertial and $\mathcal{B}$ frame, and $g$ is the acceleration due to gravity.
Body torques (from thrust differentials) are then regulated by a high-bandwidth inner loop so that commanded rotations can be maintained.
The non-linear system can then be written as a feedforward-linearized system of equations,
\begin{equation}
    \begin{split}
        \dot{\boldsymbol{x}} &= A\boldsymbol{x} + B\boldsymbol{u},\\
        y &= C\boldsymbol{x}
    \end{split}
    \quad
    \begin{split}
        \begingroup
        \setlength\arraycolsep{2pt}
        A = \begin{pmatrix}
                \zero{3}{3} & \iden{3}{3} & \zero{3}{1} \\
                \zero{3}{3} & \zero{3}{3} & \zero{3}{1} \\
                \zero{1}{3} & \zero{1}{3} & 0 \\
            \end{pmatrix}, 
        B = \begin{pmatrix}
            \zero{3}{3} & \zero{3}{1} \\
            \iden{3}{3} & \zero{3}{1} \\
            \zero{1}{3} & 1
            \end{pmatrix},
        C = \mathbb{I}_{7 \times 7},
  \endgroup
    \end{split}
    \label{eq:lin-model}
\end{equation}

with
$\boldsymbol{x}(t) \equiv [\vectorvar{p}, \vectorvar{v}, \psi]^\top$ as the state-vector and
$\boldsymbol{u}(t) = [a_n, a_e, a_d, \dot{\psi}]^\top
\equiv [\vectorvar{a}, \dot{\psi}]^\top$ 
as the feedback control input.
$\boldsymbol{u}$ is then mapped to the inner-loop command,
$\boldsymbol{u}_{\mathrm{ap}} = [R^\mathcal{B}_\mathcal{E}, T]^\top$
using a non-linear inversion map on the $2^\mathrm{nd}$-order dynamics.
We compute $\boldsymbol{u} = -K_{\mathrm{lqr}}\boldsymbol{x}$ with an infinite-horizon linear quadratic regulator (LQR).

When flying in close proximity, the propeller downwash from \Alpha{} will produce turbulent forces that act on \Bravo{}.
We model these as a lumped force, $\fext \in \R{3}$ that acts as
$\dot{\boldsymbol{x}} = A\boldsymbol{x} + B\boldsymbol{u} + \bar{B}\fext$,
with $B = \big(\begin{smallmatrix}
  \Bar{B} & 0\\
  0 & 1
\end{smallmatrix}\big)$.
Note that we ignore yaw torques, and implicitly factor all moments into their induced lateral effects.
A nominal controller for the system in Eq~\ref{eq:lin-model} regulates the state onto some reference trajectory, $\boldsymbol{x}_r(t)$.
However, it always evolves according to Eq~\ref{eq:lin-model}, and thus, will have significant tracking residuals when $\fext$ is present.

\textbf{Learning Downwash Function.}
We \textit{estimate} ground truth labels
$\hat{\vectorvar{f}}_{\mathrm{ext}}(t)$
by flying \Alpha{} and \Bravo{} at various separations using their nominal controllers, and observing the difference\footnotemark[2] between \Bravo{}'s observed and commanded lateral accelerations, $\vectorvar{a}_{\mathrm{obs}} - \boldsymbol{u}_{[1:3]}$.
A neural network is trained to predict these labels as a function of vehicle states.
Next, we briefly cover the learning process here from our prior work \cite{smith2023so} for the sake of completeness.
\footnotetext[2]{In general, commanded and observed accelerations can differ due to several physical factors.
However, here we treat them as constant biases and offsets that exist with/without $\fext$.}

We model the true function $\fext$ for \Bravo{} as dependent on a 9-dimensional state,
$\mathbf{x}\equiv[\Delta\vectorvar{p}, \vectorvar{v}^\Al, \vectorvar{v}^B]$.
Intuitively, using relative positions ($\Delta\vectorvar{p}$) allows any model (learnt or not) to be translation invariant in \SE{3}.
In addition, for docking, using absolute velocities allows us to distinguish between cases where relative velocities are zero because the vehicles are hovering, or because \Bravo{} is holding station.
Now, a na\"ive neural network architecture for learning $\fext : \R{9} \mapsto \R{3}$ can be highly inefficient, since (i) we must uniformly sample in $\R{9}$, and, (ii), the model is unaware of any symmetries and treats each data point as unique.

Therefore, we introduce a feature mapping $h: \mathbb{R}^{9} \rightarrow \mathbb{R}^6$ that preserves equivariance of $\fext$ over $\SO{2}$, and allows a network to exploit rotational symmetries:
{\small
\begin{align}\label{eq:featuremap}
    h(\mathbf{x}) =& \bigg(\!
        \frac{ \proj{\mathcal{S}_{\Al}}(\Delta \vectorvar{p} )^{\top}
                \proj{\mathcal{S}_{\Al}}(\vectorvar{v}^{\Br} )
            }
            {\| \proj{\mathcal{S}_{\Al}}(\Delta \vectorvar{p})\|_2 
                \|\proj{\mathcal{S}_{\Al}}(\vectorvar{v}^{\Br} )\|_2
            }, 
        \| \proj{\mathcal{S}_{\Al}}\left(\Delta \vectorvar{p} \right)\|_2,\nonumber\\
    &   \| \proj{\mathcal{S}_{\Al}}(\vectorvar{v}^{\Br})\|_2, 
        \left[R_{\mathcal{E}}^{\Al} \Delta \vectorvar{p}\right]_3, 
        \left[R_{\mathcal{E}}^{\Al} \vectorvar{v}^{\Br} \right]_3, \| \proj{\mathcal{S}_{\Al}}(\vectorvar{v}^{\Al})\|_2 \!\bigg),
\end{align}%
}
with
$\proj{\mathcal{S}_{\Al}}(\cdot)$ as the orthogonal projection operator onto the subspace $\mathcal{S}_{\Al} = \text{span}\{a_1, a_2\}$ (see Fig.~\ref{fig:illustration}), and
$[\cdot]_3$ denoting the third component of a vector.
Here, $h$ only provides an invariant transform over the action of \SO{2}.
For equivariance, we use the angle, $\varphi(\mathbf{x})$, between $\proj{\Al}(\Delta \vectorvar{p})$ and the positive $a_1$ axis in $\mathcal{S}_{\Al}$.
The network uses this smaller space to learn $\fext$.

\textbf{Curriculum for learning.}
Since flying under \Alpha{} can be dangerous (or even infeasible) without a downwash model in the first place,
we adopt a curriculum approach for capturing this data.
\Bravo{}'s training is divided into multiple ``stages", with \textsf{stage-0} beginning at a large vertical separation ($1.8-\SI{1.5}{m}$).
A model is trained for this stage and deployed online to collect new training samples at a smaller separation.
We repeat this in increments of \SI{0.3}{m} until \Bravo{} can collect data at $\approx$\SI{0.5}{m} (roughly 2 body-lengths of \Alpha{}).

\textbf{Docking Mechanism and Planning.}
The docking platform towed by \Alpha{} is a cylindrical bar suspended by two cables of equal length $d_p$.
It is allowed to swing and yaw (constrained by cable length) as \Alpha{} moves.
The choice of material for the cable is generally driven by its structural properties, most prominently, its weight and stiffness.
We prefer minimizing the cable's total mass to reduce the total weight that will be towed in the air.
Furthermore, a cable with a very low stiffness (high `springiness') poses operational risk to both vehicles due to induced vibrations and oscillations.
Conversely, a highly rigid cable presents itself as a large moment arm acting on \Alpha{}.
We therefore choose a cable with a balanced stiffness quotient.
In Section~\ref{sec:results} we report on our study against different cable lengths, and its effect on the magnitude of the downwash effect.

Our gripper system is a single-point gripper\footnotemark[3] that clasps the bar upon contact, and/or, when commanded by software.
The mechanism has a rotary `gate' that, when closed, can retain the bar clasped.
Similar gripper designs have been successful for docking~\cite{shankar2021multirotor}, with the key difference here being miniaturization for our smaller platform, and a faster response time to account for faster dynamics.

\footnotetext[3]{Single-point grippers are generally less sensitive to small yaw misalignments.}

\section{Experiments and Results}
\label{sec:results}
Evaluations of our learning pipeline can be found in our prior work~\cite{smith2023so}.
Here, we focus on the \textit{deployment} of the learnt model online and its effect on \Bravo{}'s docking maneuvers.
Critically, docking with \Alpha{} presents a `make-or-break' scenario, where insufficient accuracy in tracking causes a direct mission failure.

\subsection{Technical Setup}
In all our experiments (including training), \Alpha{} and \Bravo{} are quadcopters that are designed in-house using a mix of commercial off-the-shelf and specialized hardware.
They span $\approx\SI{0.27}{m}$ on the longest diagonal, and have a stock mass of $\approx\SI{0.7}{kg}$ (including batteries).
The gripper system is designed as a servo-controlled rotary `gate', with a \SI{6.35}{mm} linear patch strain gauge (from Micro-Measurements) mounted on the triggering arm.
The patch has a resistance of \SI{350}{\ohm} with a $\pm\SI{5}{\percent}$ strain range, which is sufficient to detect a contact with the docking platform.
It is calibrated in a wheatstone bridge configuration and interfaced over a serial line with an onboard Raspberry Pi (Model 4B) using an HX711 amplifier.
At the beginning of each flight, the strain gauge undergoes a calibration sequence in order to estimate its resting bias.

All the estimation, control, planning and communication algorithms run entirely onboard the Raspberry Pi.
The flight controller uses a slightly customized ArduCopter firmware, while the high-level control and planning is done using Freyja~\cite{shankar2021freyja} running over ROS2.
For our indoor experiments, a motion-capture system provides positioning data independently to both vehicles, and they communicate over an underlying network.
Note that our architecture is easily `transported' to the field (where the vehicles can receive their independent state data using GNSS), and does not depend on centralized systems.

\subsection{Studying Effects of Cable Length}
Docking with a reduced cable length is a key innovation in this work.
Shorter cables can generally be preferred due to a variety of reasons: they offer more stiffness (less swinging), are easier for \Alpha{} to control, and pose smaller operational risks.
However, in the absence of a downwash model, \Bravo{} will generally prefer longer cables to reduce aerodynamic disturbances acting on it.
In this subsection, we preform a study on this effect over different cable lengths, and evaluate \Bravo{}'s docking performance at these vertical offsets.
Note that longer cables are also likely to swing more, thus offering less contact force for the sensor on the gripper (causing more false negatives).

\begin{figure}[t]
    \centering
    \includegraphics[width=0.49\textwidth]{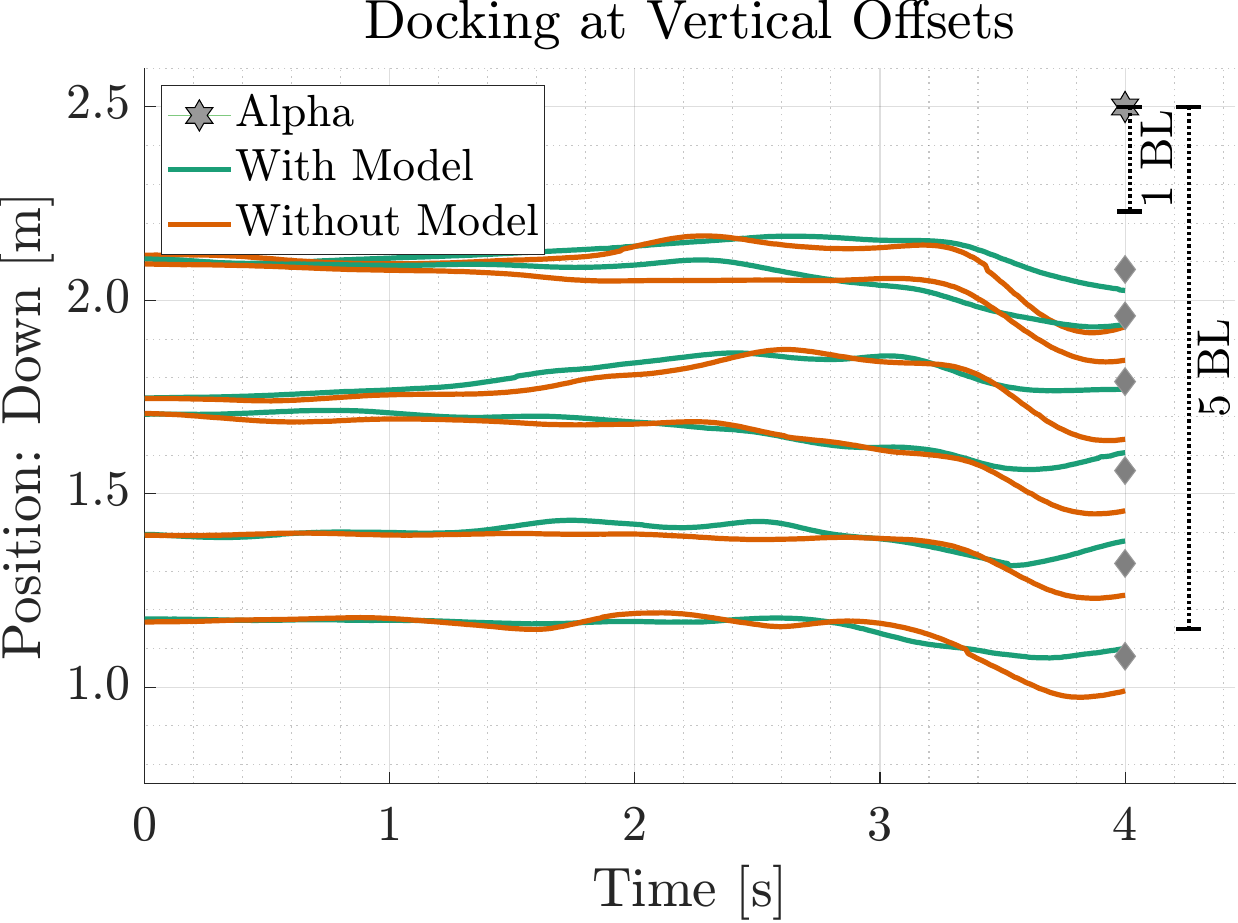}
    \includegraphics[width=0.49\textwidth]{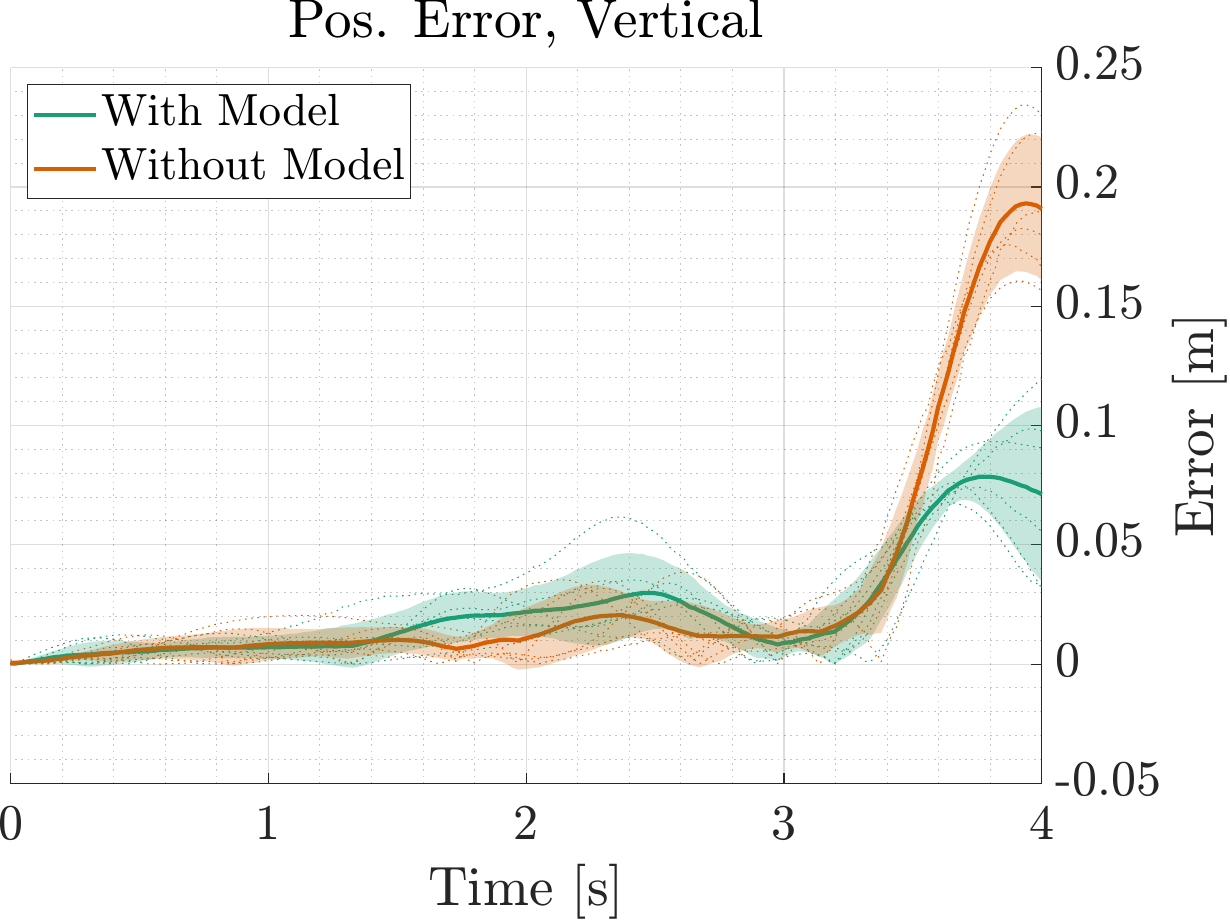}
    \caption{(Left) Trajectories from docking attempts at various vertical offsets from \textit{Alpha} (located at altitude \SI{2.5}{m}). (Right) Combined vertical positioning error with one standard deviation in shaded regions. The final docking happens right before the \SI{4}{s} mark.}
    \label{fig:vsep-tests}
\end{figure}

\begin{table}[b]
\renewcommand{\arraystretch}{1.1}
\centering
\caption{Statistics integrated over the final \SI{0.5}{s} of the mission.}
\label{tab:offsets-table}
\begin{tabular}{ccccc|cccc}
                            & \multicolumn{4}{c|}{\textbf{Without Model}} & \multicolumn{4}{c}{\textbf{With Model}} \\
                            \cline{2-9} 
\multicolumn{1}{c|}{Offset\newline[m]} &  
                                \begin{tabular}[c]{@{}c@{}}Error\\Down [m]\end{tabular} & 
                                \begin{tabular}[c]{@{}c@{}}Error\\3D [m]\end{tabular} & 
                                \begin{tabular}[c]{@{}c@{}}Predicted\\Force [\SI{}{m/s^2}]\end{tabular}&
                                Result &
                                \begin{tabular}[c]{@{}c@{}}Error\\Down [m]\end{tabular} & 
                                \begin{tabular}[c]{@{}c@{}}Error\\3D [m]\end{tabular} &
                                \begin{tabular}[c]{@{}c@{}}Predicted\\Force [\SI{}{m/s^2}]\end{tabular} &
                                Result \\
                                \cline{1-9} 
\multicolumn{1}{c|}{0.36} & 2.28 & 2.38 & N/A & Miss & 1.06 & 1.21 & 6.13 & Dock \\
\multicolumn{1}{c|}{0.47} & 1.72 & 1.82 & N/A & Miss & 1.02 & 1.59 & 2.84 & Dock \\
\multicolumn{1}{c|}{0.64} & 2.08 & 2.10 & N/A & Miss & 1.09 & 1.70 & 2.21 & Dock \\
\multicolumn{1}{c|}{0.87} & 1.85 & 1.89 & N/A & Miss & 0.70 & 1.11 & 4.72 & Dock \\
\multicolumn{1}{c|}{1.11} & 1.67 & 1.86 & N/A & Miss & 0.77 & 1.73 & 2.49 & Dock \\
\multicolumn{1}{c|}{1.35} & 1.90 & 1.93 & N/A & Miss & 0.85 & 1.19 & 2.24 & Dock \\ \cline{1-9}
\end{tabular}%

\end{table}

For these tests, we affix \Alpha{} on a rigidly mounted stand, and command it a hover thrust.
The docking platform is suspended directly underneath it with an adjustable vertical offset of the cable.
We sample six cable lengths that lie between one and five body-lengths of the vehicles.
\Bravo{} is initialized at the same location laterally (the vertical position is adjusted according to the cable offset), and repeats its docking missions with and without the model deployed.
Note that these tests are realistic, even if some of the conditions are held ideal.
For instance, keeping \Alpha{} rigidly attached helps shield the experiment from its positioning errors when long cables are used (the flight may even be infeasible in extreme cases).
\Bravo{} executes missions as though \Alpha{} were in free flight.

Fig.~\ref{fig:vsep-tests}(left) shows vertical positioning results from these evaluations.
Unsurprisingly, we observe that the differences with or without the model become most evident towards the end of \Bravo{}'s missions ($\approx\SI{4}{s}$, when it is most directly underneath \Alpha{}).
The mean and standard deviation for vertical positioning errors in all these tests are shown in Fig.~\ref{fig:vsep-tests}(right).
We observe a surprising consistency in errors with/without the model at all vertical offsets (cable lengths) in our test, indicating that even at 5 body-lengths (BL), \Bravo{} is still likely to experience noticeable disturbance from \Alpha{}.
As also noted in Table~\ref{tab:offsets-table}, with model compensations enabled, the system is able to effect a docking in each of these cases, while it fails consistently without it.
Since the most prominent effects appear to be concentrated in the terminal \SI{0.5}{s}, the table reports some key statistics integrated over this period, i.e.,
$\int_{t-0.5}^{t} (\cdot) dt$.
Most prominently, we see the large magnitude of corrections produced by the model at the closest offset, after which it trails off quite sharply.
Also, we notice that the vertical positioning error plateaus in both cases after about 3 body-lengths.
However, the largest positioning error using the model corrections remains much smaller than the smallest positioning error without any model -- and this is key to achieving a physical dock at any separation.

\subsection{Docking Mid-Air}
Next, we perform experiments with \Alpha{} in free-flying stationary hover, with a cable-suspended docking platform at a fixed vertical offset.
We choose the smallest vertical offset ($\approx$\SI{0.36}{m}) from Table~\ref{tab:offsets-table}, which corresponds to a little over one body-length of our platforms.
Note that at this proximity, we must additionally account for vertical protrusions such as GPS mounts and batteries on either vehicle.
\Bravo{} can communicate with \Alpha{} in order to exchange key pieces of information, such as docking status, and obtaining \Alpha{}'s relative state.
As before, this traffic is routed through an infrastructure network, but the vehicles do not have a centralized controller or manager.

In our evaluations, \Bravo{} is initialized at different starting locations, and plans a smooth parametric trajectory over the horizon $t_h$ that positions its gripper at a predicted [future] location of the docking platform.
The initial state includes small randomizations in order to create different profiles for the approach trajectories.
We use $d_p = \SI{0.36}{m}$ and $\epsilon_g = \SI{0.1}{m}$ (the latter corresponds to the `aperture' of our gripper's gate).
For simplicity, we assume that the airspace has no obstacles or other participants.

\begin{figure}
\centering
    \begin{subfigure}[b]{0.48\textwidth}
        \includegraphics[width=\textwidth]{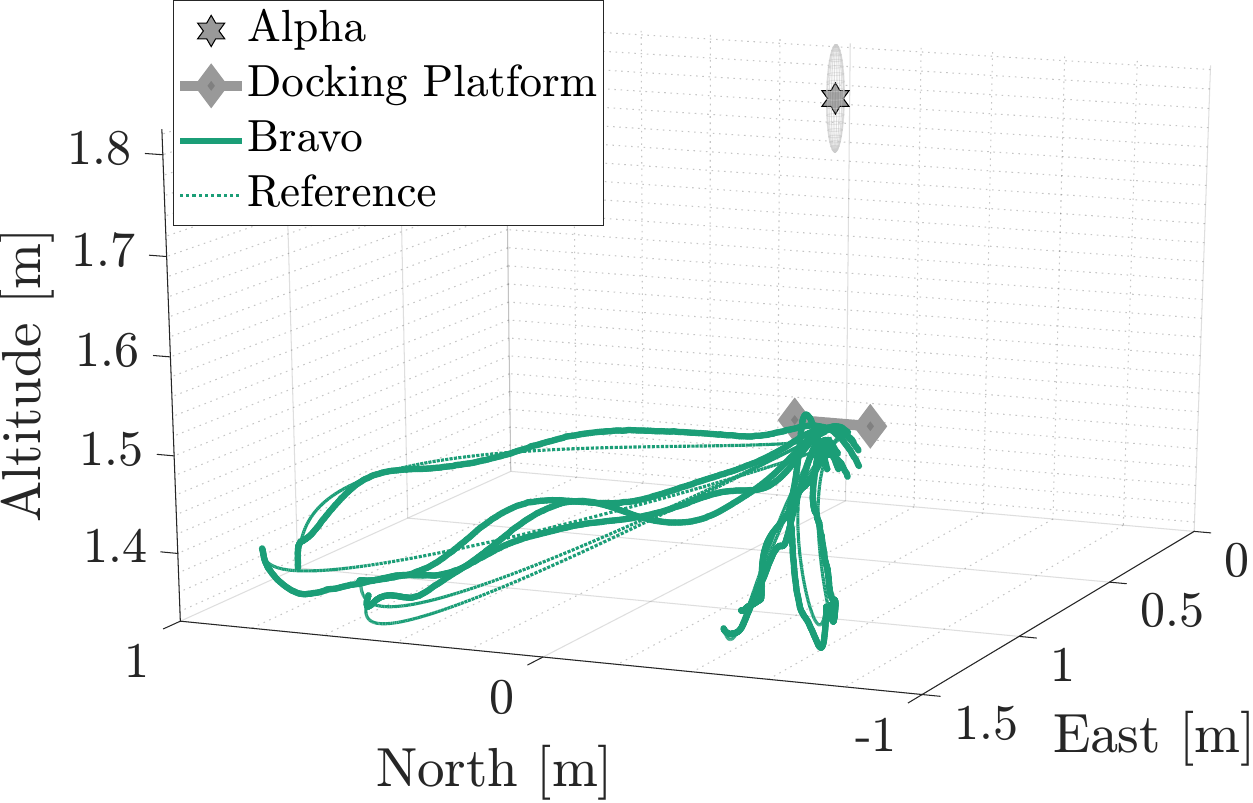}
        \caption{}
         \label{fig:st-left}
    \end{subfigure}
    \hfill
    \begin{subfigure}[b]{0.48\textwidth}
        \includegraphics[width=\textwidth]{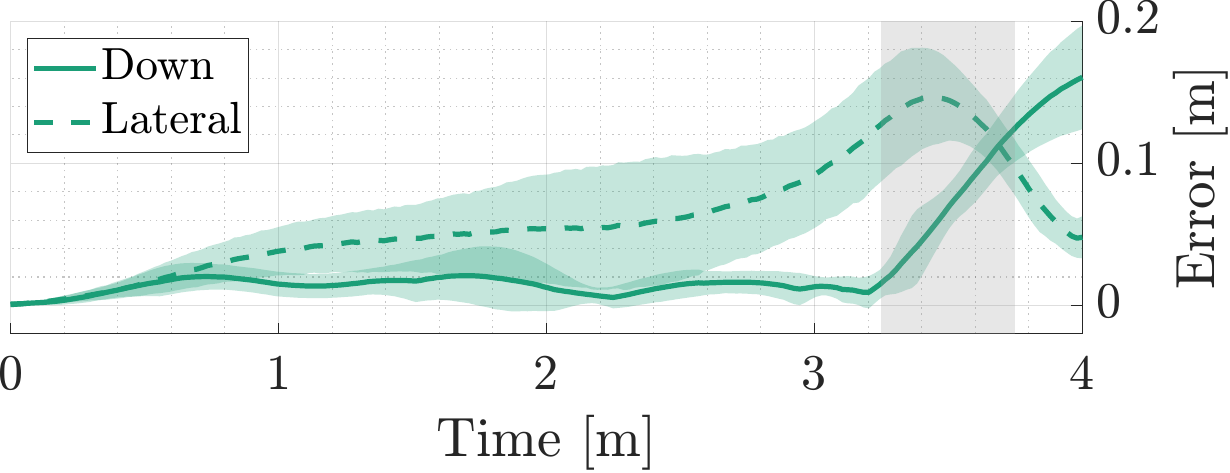}
        \includegraphics[width=\textwidth]{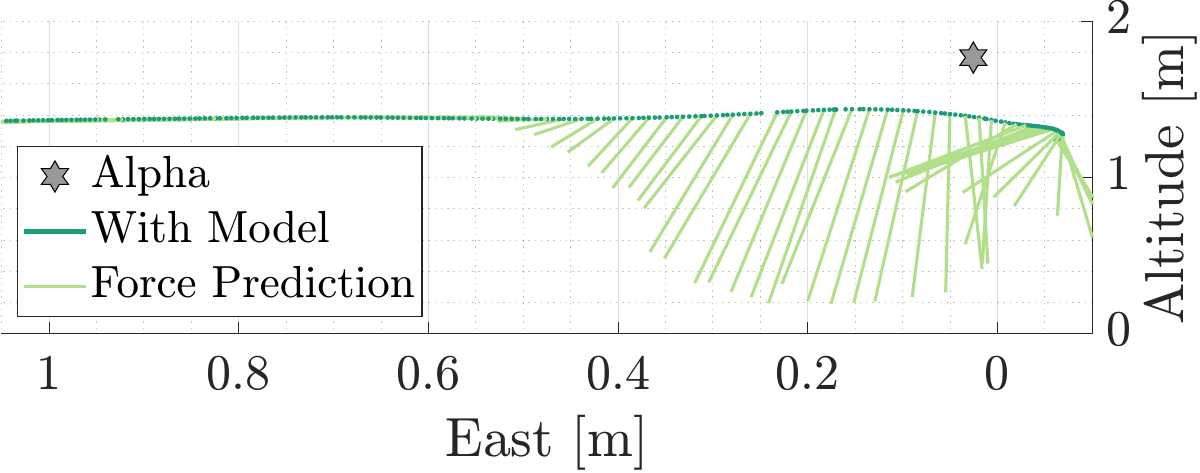}
        \caption{}
        \label{fig:st-right}
    \end{subfigure}
    \caption{Evaluations with \Alpha{} hovering and \Bravo{} commanded to execute a docking maneuver towards the estimated target location (of the platform). The gray shaded area in (b) represents the interval where docking is likely to happen.}
    \label{fig:stationary-evals}
\end{figure}

Fig.~\ref{fig:stationary-evals} shows 10 evaluations of \Bravo{}'s docking maneuver initialized at four different locations and states, with \Alpha{} hovering at a fixed location.
Note that since \Alpha{} is not rigidly affixed, it exhibits tracking errors and uncertainty in positioning (one standard deviation is shown as a shaded ellipsoid around its location in Figure~\ref{fig:st-left}).
In each instance, \Bravo{} plans a $\approx\SI{4}{s}$-long trajectory that positions its gripper exactly $d_p = \SI{0.36}{m}$ distance below \Alpha{} .
We see the effect of deploying the model in Fig.~\ref{fig:st-left}/\ref{fig:st-right}: \Bravo{} is successful in effecting a final dock repeatably in 8 out of 10 instances.

Due to the dynamic motion of both vehicles, the actual physical dock happens slightly before \SI{4}{s}, shown in the shaded region in Fig.~\ref{fig:st-right}(top).
Fig.~\ref{fig:st-right}(below) shows a planar view of a successful instance with model's force prediction vectors overlaid (these are the compensations incorporated by the controller).
As noted before with the static setup, these increase in magnitude quite sharply as \Bravo{} approaches close to its terminus point.
They contribute most significantly in the vertical axis, and are thus critical for maintaining accurate relative altitude.

\begin{wrapfigure}{R}{0.48\textwidth}
    \centering
    \includegraphics[width=0.48\textwidth]{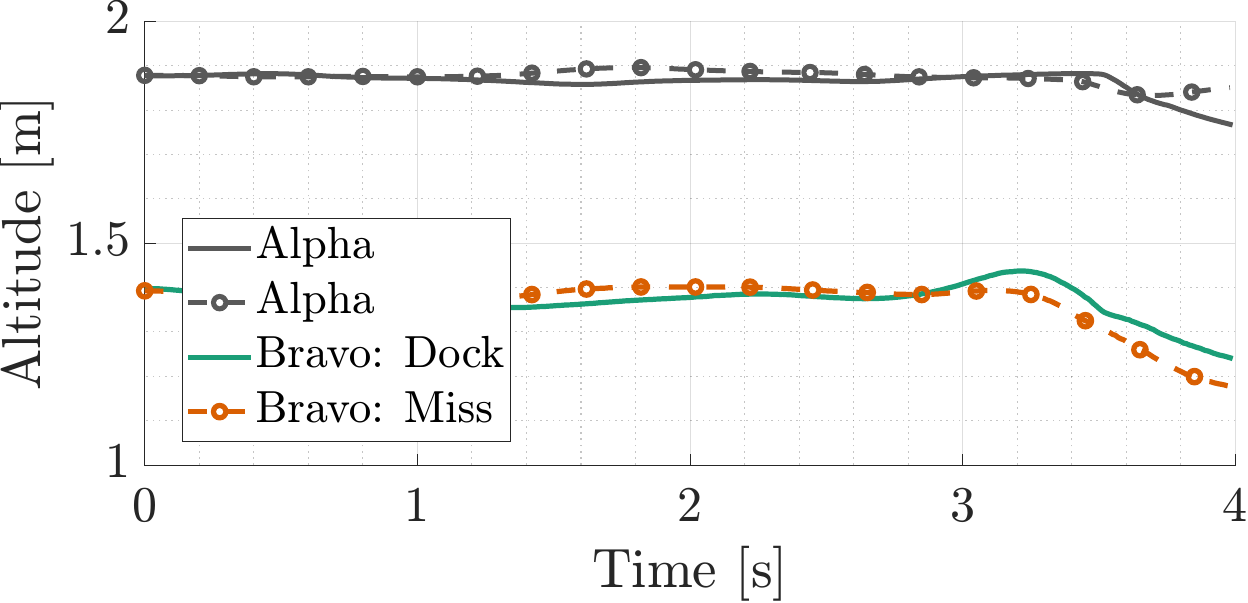}
    \caption{Altitude tracking during a successful and missed attempt. In the successful case, both \Bravo{} and \Alpha{} lose altitude after docking, while in a missed attempt, only \Bravo{} does.}
    \vspace{-0ex}
    \label{fig:sample-eval}
\end{wrapfigure}
The average vertical tracking error during this final docking period is typically less than \SI{0.1}{m}.
This is similar to what we observed before with the static setup, however, there is a key difference at the docking instant.
We observe that unlike in Fig.~\ref{fig:vsep-tests}, the vertical error appears to be increasing at the \SI{4}{s} mark.
This is due to the fact that once the two vehicles are docked, \Bravo{} reduces its thrust collective, while \Alpha{} must now carry its weight.
This momentary transition period causes \Alpha{} to dip, which appears as an increased positioning error for \Bravo{}.
We also note that the converse is true for lateral errors: once the two vehicles are docked, their relative positioning error decreases.
Fig.~\ref{fig:sample-eval} shows altitude tracking for both vehicles during a successful and a missed attempt.
We see that when successful, both \Alpha{} and \Bravo{} lose altitude once docked, whereas in a missed attempt, only \Bravo{}'s loses altitude while \Alpha{} continues to hover (with a minor dip).

\begin{wrapfigure}{R}{0.48\textwidth}
    \centering
    \includegraphics[width=0.48\textwidth]{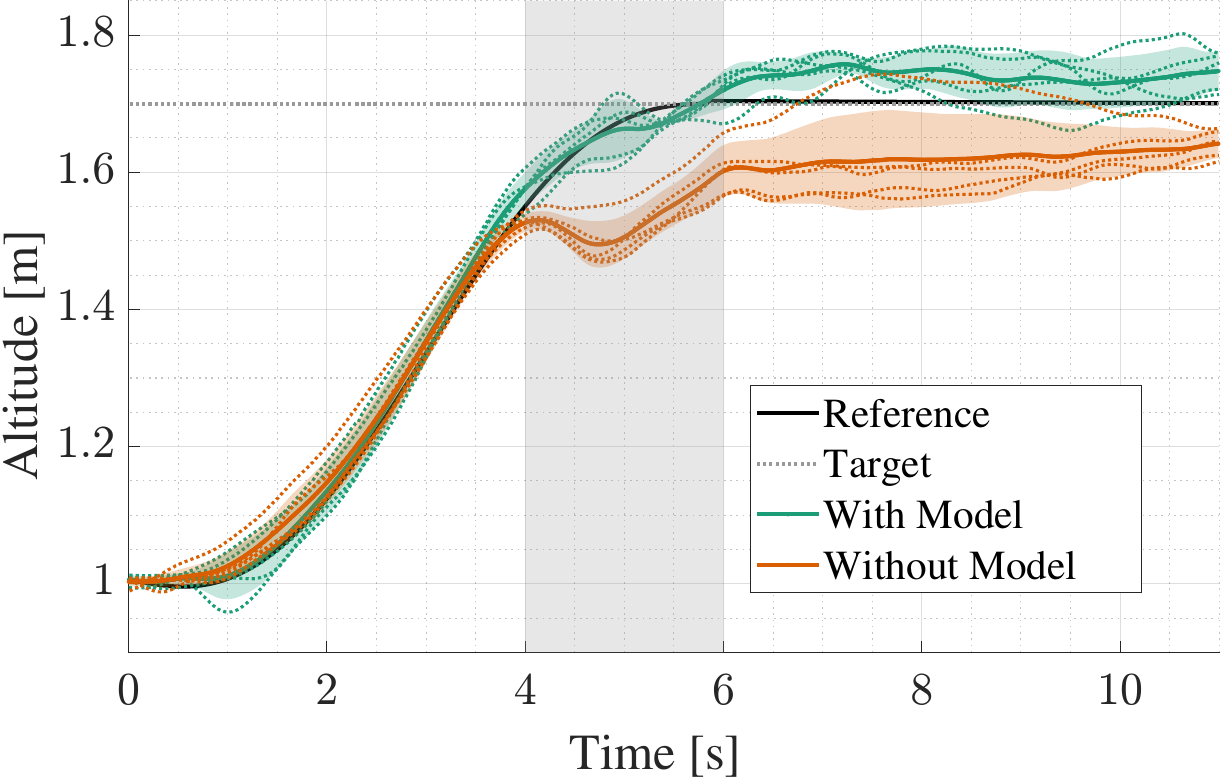}
    \vspace{-4ex}
    \caption{Evaluations with \Alpha{} in motion. \Bravo{} holds formation with \Alpha{} for \SI{5}{s} after the planned docking point (at $\approx\SI{6}{s}$).}
    \vspace{-0ex}
    \label{fig:dynamic-evals}
\end{wrapfigure}

\textbf{Alpha in Motion.}
Our downwash model is trained to handle cases where \Alpha{} is in near-hover.
Nevertheless, we also study the case where \Alpha{} is commanded to fly at a fixed horizontal speed (\SI{0.25}{m/s}), while \Bravo{} plans a trajectory that first meets the target at \SI{6}{s}, and then holds formation for the subsequent \SI{5}{s}.
Our tests had very limited success in this scenario, usually attributed to \Bravo{}'s oscillatory behavior that prevents the servo actuation from clasping the docking bar (see Section~\ref{sec:discussion} for more discussion).

Fig.~\ref{fig:dynamic-evals} shows a temporal view of \Bravo{}'s altitude reference tracking for 10 evaluations (the target altitude
is aligned to \Alpha{}'s trajectory).
Again, we note that when \Bravo{} first enters \Alpha{}'s downwash zone ($4-\SI{6}{s}$), its performance dips dramatically without model compensations.
The mean error (solid curves) in these instances is over \SI{0.2}{m}, nearly 1 body-length for our vehicles.
In contrast, with the model compensations applied, the error is reduced to less than \SI{0.05}{m}, and is nearly zero at \SI{6}{s}.

\section{Discussion and Conclusion}
\label{sec:discussion}
Our evaluations with \Alpha{} in hover are largely successful, with a failed instance usually attributed to the limits on physical actuation.
The gripping mechanism is triggered by a contact-force sensing element, which has a finite non-zero response time characteristic.
Additionally, we estimate a digital latency of approximately \SI{15}{ms} in recognizing a contact in the interfacing software.
Filtering and inter-process communication induce some delays, and the servo actuation has its own response time (governed by the torque-curve of the servo).
Consequently, in a missed docking attempt, the system recognizes a contact, the servo actuation is triggered, but the docking platform ``slips through" the mechanism before it is clasped securely.

In the tests where \Alpha{} is in motion, we observe that while \Bravo{} is successful in positioning itself in proximity to \Alpha{}, its the terminal state is insufficient for effecting a physical dock.
These failures are similar to the hover case, and suggest that the aforementioned physical limitations are strongly at play.
We also see that these missions would all likely fail without our downwash model, unless the separation ($d_p$) is increased.
This is undesirable, as discussed, since long cables pose additional operational risk to \Alpha{}.

Fig.~\ref{fig:dynamic-evals} shows an interesting artifact of a feedback controller/observer (LQG):
the error diminishes slowly over the last \SI{5}{s} even without model compensation.
In practice, such missions have a strict timeout after which the approach must be aborted.
While the observer can be tuned for a more rapid decay, that comes at the expense of robustness in regular undisturbed flight.
Furthermore, in practice, the downwash function changes so rapidly that a convergence is not guaranteed.

Finally, this work has presented the first demonstrations of two multirotors docking in such close vertical proximity.
This novel application of learnt geometry-aware neural network models unlocks several other use-cases for multirotors interacting with each other as part of planned missions.
Our future work will primarily aim to develop models that are applicable for cases when \Alpha{} is moving rapidly, or has a different physical geometry.

\subsubsection{Acknowledgment.}
This work was supported in part by ARL DCIST CRA W911NF-17-2-0181, the European Research Council (ERC) Project 949940 (gAIa), and by a gift from Arm.
Jennifer Gielis also helped with coordinating multi-vehicle flights.
Their support is gratefully acknowledged.

\bibliographystyle{spmpsci_unsrt}
{\footnotesize
\bibliography{aj_refs}
}

\end{document}